\documentclass[letterpaper, 10 pt, conference]{ieeeconf}
\usepackage[utf8]{inputenc}

\IEEEoverridecommandlockouts

\usepackage{amsmath} %
\usepackage{amssymb}  %
\usepackage[caption=false, font=footnotesize]{subfig}
\usepackage{adjustbox}
\usepackage{siunitx}
\usepackage{tikz}
\usepackage{gnuplot-lua-tikz}
\usetikzlibrary{shapes,positioning,arrows,arrows.meta,math,matrix,calc,patterns,backgrounds,plotmarks,fit}
\usepackage{algorithm}
\usepackage[noend]{algorithmic}
\usepackage{hyperref}
\usepackage{bm}
\usepackage[normalem]{ulem} %

\newcommand{\vdotssm}{\vbox{\baselineskip=2pt \lineskiplimit=0pt \kern1pt \hbox{.}\hbox{.}\hbox{.}}}

\renewcommand{\vec}[1]{\mathbf{#1}}
\newcommand{\mat}[1]{\mathbf{#1}}

\newcommand{\Tau}{\mathcal{T}}

\DeclareMathOperator*{\argmin}{argmin}

\newcommand{\JI}[1]{#1}
\newcommand{\st}[1]{}

\title{\LARGE \bf
GOMP: Grasp-Optimized Motion Planning for Bin Picking}

\author{Jeffrey Ichnowski$^{*}$, Michael Danielczuk$^{*}$, Jingyi Xu$^{*}$, Vishal Satish$^{*}$, and Ken Goldberg$^{*}$%
\thanks{$^{*}$Jeffrey Ichnowski, Mike Danielczuk, Jingyi Xu, Vishal Satish, and Ken Goldberg are with the AUTOLAB, University of California at Berkeley,
Berkeley, CA 94720, U.S.A.
        {\tt\small \{jeffi, mdanielczuk, jingyi\_xu, vsatish, goldberg\}@berkeley.edu}}%
}

\begin{document}

\maketitle
\thispagestyle{empty}
\pagestyle{empty}

\begin{abstract}
Rapid and reliable robot bin picking is a critical challenge in automating warehouses, often measured in picks-per-hour (PPH). We explore increasing PPH using faster motions based on optimizing over a set of candidate grasps. The source of this set of grasps is two-fold: (1) grasp-analysis tools such as Dex-Net generate multiple candidate grasps, and (2) each of these grasps has a degree of freedom about which a robot gripper can rotate. In this paper, we present Grasp-Optimized Motion Planning (GOMP), an algorithm that speeds up the execution of a bin-picking robot's operations by incorporating robot dynamics and a set of candidate grasps produced by a grasp planner into an optimizing motion planner. We compute motions by optimizing with sequential quadratic programming (SQP) and iteratively updating trust regions to account for the non-convex nature of the problem.  In our formulation, we constrain the motion to remain within the mechanical limits of the robot while avoiding obstacles.  We further convert the problem to a time-minimization by repeatedly shorting a time horizon of a trajectory until the SQP is infeasible.  In experiments with a UR5, GOMP achieves a speedup of 9x over a baseline planner.
\end{abstract}

\section{Introduction}
With increasing e-commerce, robots are in demand for bin picking.
This ``pick-and-place'' operation is ostensibly simple, as humans can do it will little effort.
However, the same is not true for robots.
Having robots take on this task requires tackling many computational challenges in order to make it reliable and fast, including sensing, grasp analysis, motion planning, and executing the actual robot arm's motion.
With recent advances speeding up sensing and grasp analysis, motion planning and robot arm movement are rapidly becoming the bottleneck~\cite{morrison2019learning,satish2019policy}.
In this paper, we present Grasp-Optimized Motion Planner (\emph{GOMP}), an algorithm that speeds up the execution of a bin-picking robot's operations by incorporating the dynamics of a robot arm and a \emph{set} of candidate grasps produced by grasp analysis into an \emph{optimizing motion planner}.

The pipeline for GOMP starts with the observation that grasp analysis tools can provide a set of candidate grasps.
A motion planner can then select a grasp from the set that leads to the fastest motion between picking up and placing the object.
The variation in speed of motion comes from both location and angle of grasp.
Grasping locations and angles closer to the placement configuration, or that present alternate routes around obstacles can allow for faster bin picking motions.

When generating optimized trajectories between pick and place, GOMP allows the grasp point to vary over one degree of freedom.
This degree of freedom for parallel-jaw grippers is around the grasp axis.
For suction, the degree of freedom is around the suction cup contact normal.
While traditional approaches limit the grasp to a single direction of approach (e.g., top-down) to a grasp-analysis frame, we propose allowing an optimization to select an approach angle that results in the same pair of points in contact with the gripper jaws, while allowing the robot to make faster motions.

\begin{figure}[t]
    \centering
    \includegraphics[width=3.4in,trim={346 54 366 46},clip]{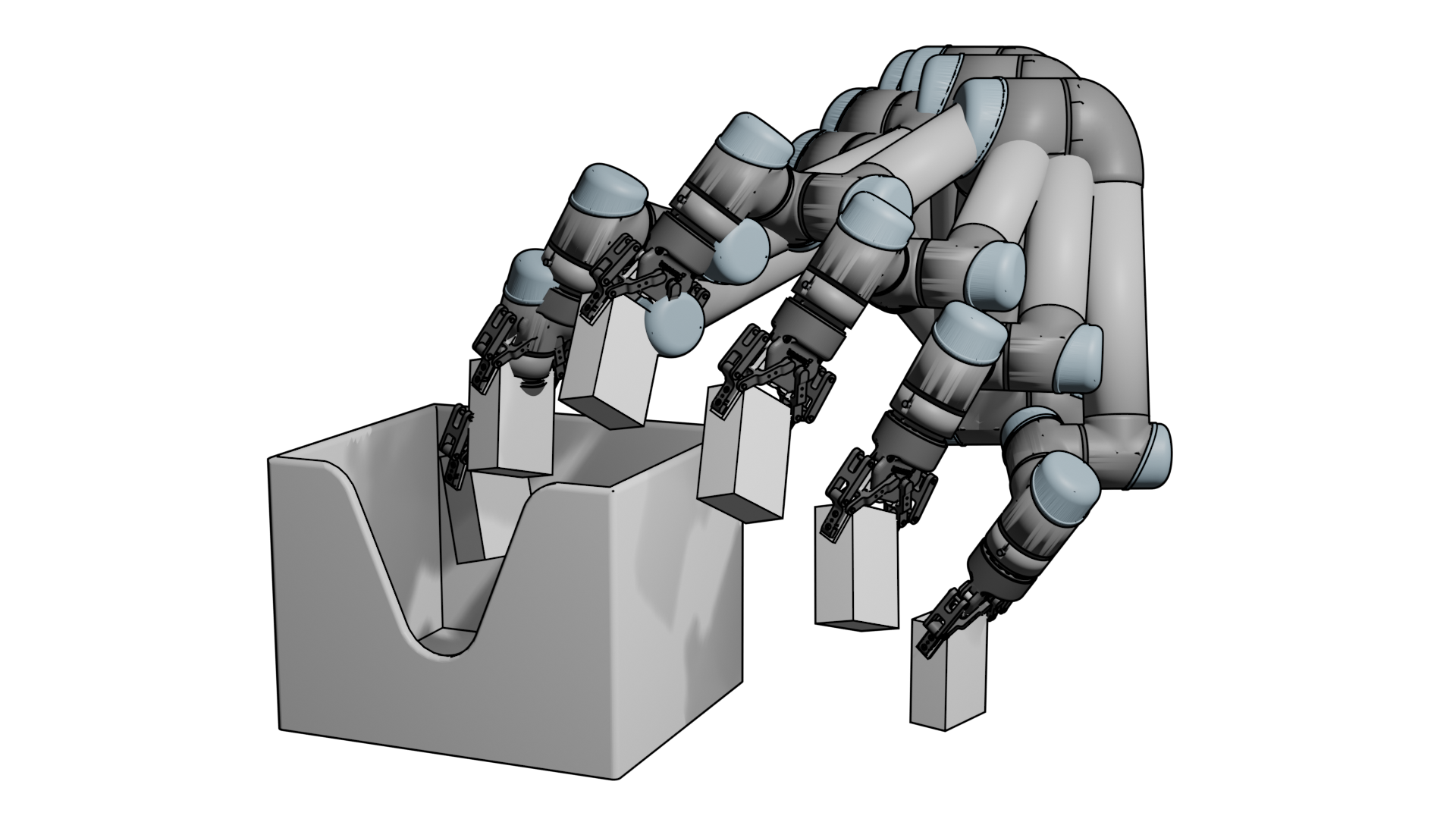}
    \caption{GOMP computes minimum-time trajectories that avoid obstacles, while constraining the start and end configurations to a \emph{set} computed by grasp analysis. Although many bin-picking systems limit the robot to picking from a top-down vector, the start and end configurations of a trajectory generated by GOMP are allowed to vary over the degree of freedom as defined by the grasp axis. This figure shows the optimal trajectory computed by GOMP for the UR5 robot with obstacles as illustrated.}
    \label{fig:gomp_motion}
\end{figure}

GOMP incorporates the mechanical limits (e.g., each joint's maximum torque and velocity) of the robot arm to minimize the resulting motions' execution time.
This minimization is non-convex, as it must avoid obstacles in the robot's workspace (e.g., the bin sides) and allow for degrees of freedom in the forward-kinematic space of the start and goal poses while taking into account the mechanical limits.
To address this problem, GOMP uses a sequential convex optimization on a discretization of a trajectory into a fixed number of waypoints at a fixed time interval that:
(1) iteratively updates trust regions between convex optimization steps to account for the constraints of the start and goal poses and obstacles,
(2) explicitly constrains joint motions to remain within their velocity and torque limits, and
(3) minimizes the acceleration to produce smooth trajectories.

To minimize motion time, the sequential convex optimization above is repeated with a decreasing number of waypoints until the minimization is no longer feasible.
This repeated optimization is sped up through the use of warm starting the subsequent optimization based upon the previous optimization.

In experiments, we integrate grasp analysis with the optimizing motion planner and deploy it on a UR5 robot arm~\cite{UR5robot} with a Robotiq gripper~\cite{Robotiq2f85}.
For the grasp analysis, we use a modified Dex-Net 4.0~\cite{mahler2019learning} which can produce multiple grasp candidates in sub-second times.
The result of these is an optimized motion with no observed loss of reliability, resulting in faster PPH than prior work.

This paper provides 3 main contributions:
\begin{enumerate}
    \item The GOMP problem formalization and algorithm for minimizing the execution time of a robot's motion while accounting for obstacles, robot's mechanical limits, and grasp constraints.
    \item Grasp-analysis-based pose constraints that allow an optimization to speed up motions by varying over a degree of freedom implied by the gripper's design.
    \item Experimental results suggesting GOMP motions can significantly speed up the bin picking pipeline.
\end{enumerate}

\begin{figure}[t]
  \centering
  \begin{tabular}{@{}c@{\;}c@{\;}c@{}}
    \subfloat[top-down]{\includegraphics[height=1.5in,trim={562 108 500 0},clip]{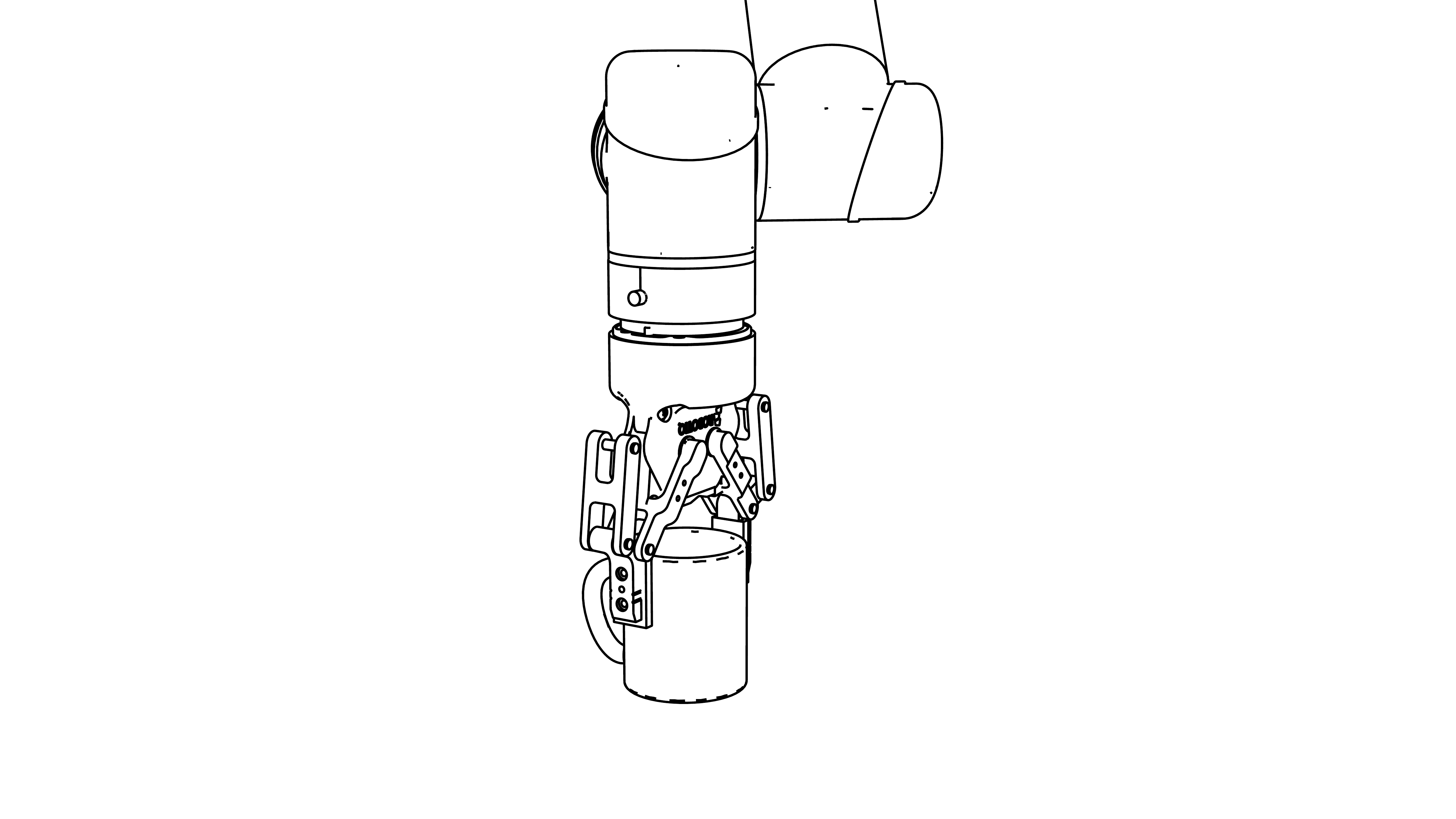}} &
    \subfloat[left \ang{60}]{\includegraphics[height=1.5in,trim={292 108 676 0},clip]{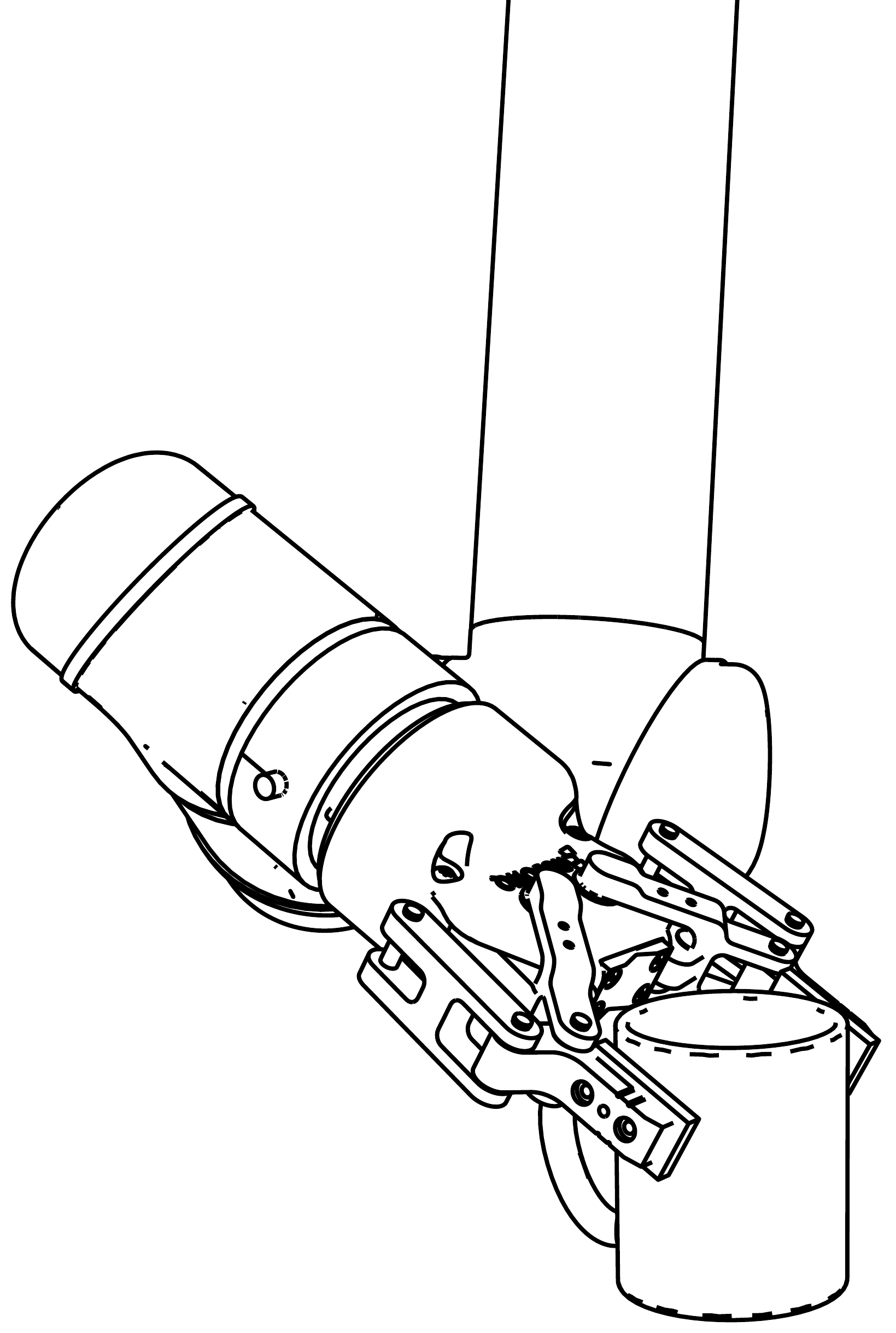}} &
    \subfloat[right \ang{60}]{\includegraphics[height=1.5in,trim={570 108 238 0},clip]{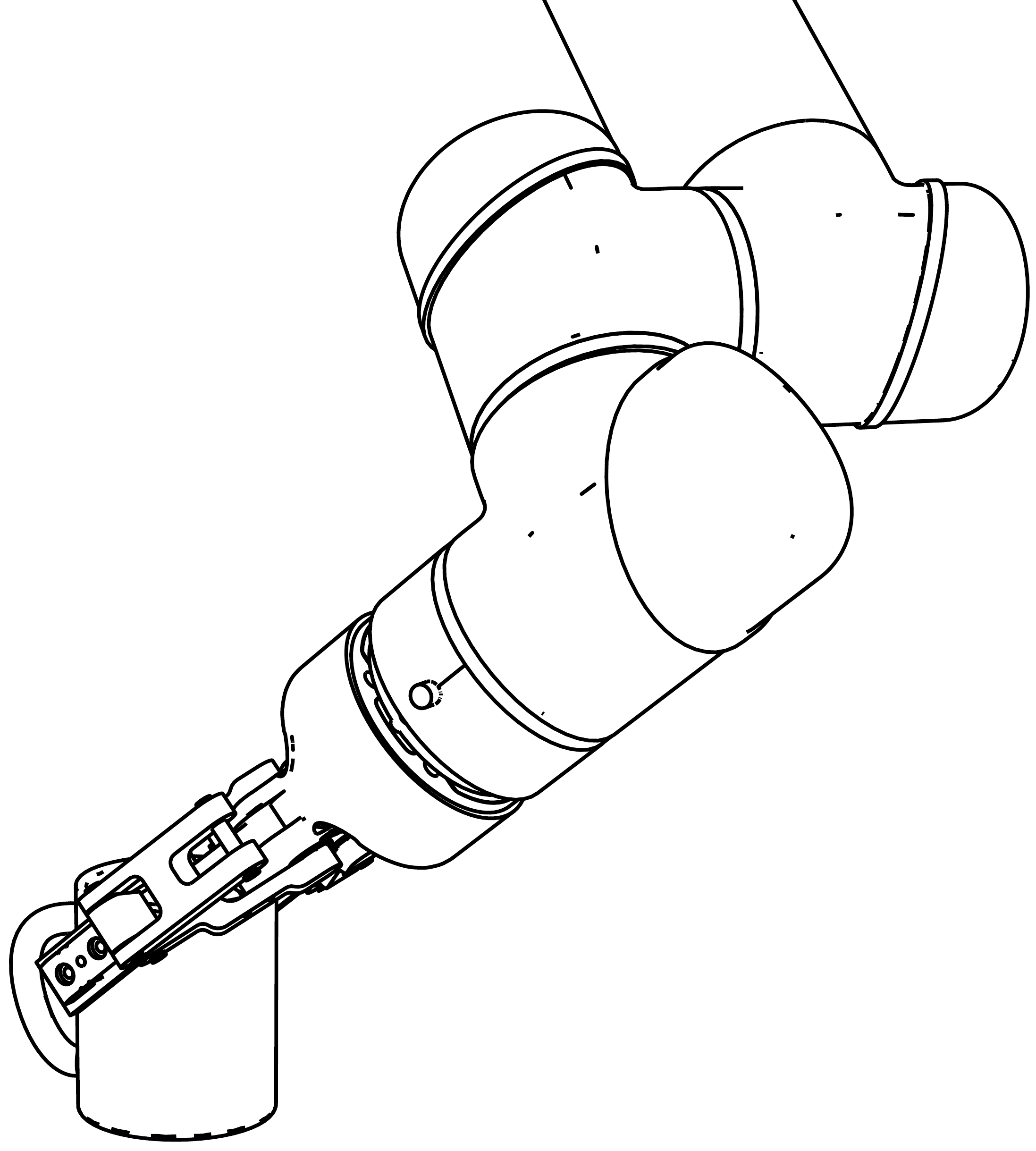}}
  \end{tabular}
  \caption{Grasp analysis produces a pair of grasp points that a parallel jaw gripper is able to grasp from any angle rotated around the grasp axis created connecting the two grasp points.  The optimizing motion planner starts with the traditional top-down grasp from (a) and optimizes around the axis to find the fastest motion.}
  \label{fig:grasp_dof}
\end{figure}

\section{Related Work}

Data-driven grasp-analysis or grasp planning algorithms for parallel jaw grippers, such as Dex-Net~\cite{mahler2016dex,mahler2017dex,mahler2017learning,mahler2019learning}, GG-CNN~\cite{morrison2019learning}, GPD~\cite{ten2017grasp}, or FC-GQ-CNN~\cite{satish2019policy}, typically take sensor input (e.g., an object mesh, a depth-camera image), perform some pre-processing (e.g., image inpainting), and produce either a grasp or grasp quality score for a presampled grasp candidate~\cite{kappler2015leveraging}. The majority of these algorithms are based on convolutional neural networks (CNNs) and may be learned from human annotations~\cite{lenz2015deep}, simulated training data~\cite{johns2016deep, viereck2017learning}, human or self-supervised labels from grasps attempted on a physical system~\cite{kalashnikov2018scalable,pinto2016supersizing,levine2016learning} or a combination of the above~\cite{bousmalis2018using}. These data-driven methods often represent grasps using a center-axis~\cite{mahler2019learning} or rectangle formulation~\cite{lenz2015deep} in the image plane, resulting in 4 degrees of freedom (a 3D translation, plus a rotation about the camera z-axis). Recent work has also explored introducing additional degrees of freedom for grasps in cluttered environments~\cite{mousavian20196,murali20196,yan2018learning,liu2019generating}, noting that top-down grasps leave out a wide range of feasible high quality grasps on many objects~\cite{eppner2019billion}. In this paper, we build upon the 4 degree of freedom representation commonly used with CNNs and propose using the output coordinate frame as a basis for an infinite set of grasps around the grasp axis. We note that this additional degree of freedom applies also to suction cup grasps, which can be rotated around the surface normal of the grasp point.   While we propose that many grasp-analysis algorithms could be incorporated into GOMP, we use FC-GQ-CNN as it can produce multiple grasp candidates in sub-second times.%

Generating motion plans from a multiple start to multiple goal configurations (e.g., by varying degrees of freedom on the configurations), is well-suited for sampling-based motion planners such as PRM~\cite{Kavraki1996_TRA}, RRG~\cite{Karaman2011_IJRR}, and bi-directional RRTs~\cite{kuffner2005efficient}.  While asymptotically-optimal variants of these planners~\cite{Karaman2011_IJRR} are guaranteed to produce optimal motions eventually, the slow convergence rate %
of these planners in 6+ degree-of-freedom robot arms does not lend itself to producing time-optimal motion plans fast enough to improve \JI{\st{M}}PPH.  While sampling-based motion planners lend themselves to parallelization~\cite{Amato1999_ICRA,Ichnowski2014_TRO}, the associated speed up in convergence rate may still be insufficient.  In any case, the generated motions often still require subsequent smoothing~\cite{pan2012collision}\JI{, \st{and/or}} time-parameterization\JI{\st{ of the motions}}~\cite{pham2014general}\JI{, or both~\cite{kunz2012time}} in order to run smoothly and quickly on a robot arm.

Optimizing motion planners for robot such as  Trajopt~\cite{schulman2013finding}, CHOMP~\cite{ratliff2009chomp}, STOMP~\cite{kalakrishnan2011stomp},
ITOMP~\cite{park2012itomp},
based on interior point optimization~\cite{kuntz2017fast}, based on gradients \cite{campana2016gradient}, and based on Gaussian Process~\cite{mukadam2018continuous}, through various formulations of a minimization problem work to produce a locally minimized trajectory from a start to goal pose while avoiding obstacles.
In prior formulations, the trajectory is discretized into a sequence of waypoints, and the constrained minimization objective is cast either as sum-of-squared distances between waypoints, or sum-of-distances between waypoints, without taking into account the mechanical limits of the robot arm (i.e., requiring a subsequent time-parameterization step).
In this paper, like prior work, we discretize the trajectory into a sequence of a fixed number of waypoints.
Unlike prior work, we minimize the sum-of-squared accelerations while incorporating mechanical limits and dynamics between waypoints (to produce smooth motions), and then repeat the process with fewer waypoints until we find a minimum-time trajectory.
With the observation that our obstacle-avoidance problem is relatively simple compared to those addressed by prior work, we use constraints similar to those of Trajopt, though with a faster implementation using a different obstacle model.

Integrated grasp and motion planning algorithms find collision-free trajectories to grasps or grasp sets that are precomputed or synthesized during the planning process.
Dragan et al.~\cite{dragan2011manipulation} use CHOMP to compute an optimized motion plan to a discretized goal set.
Wang et al.~\cite{wang2019manipulation} extend on Dragan's approach to include learning and online synthesis of grasps that increase the goal set during optimization.
Vahrenkamp et al.~\cite{vahrenkamp2010integrated} propose a planner (GraspRRT) that integrates finding a feasible grasp, inverse kinematics, and finding a collision-free motion plan using a sampling-based motion planner.
Fontanals et al.~\cite{fontanals2014integrated}
and Gravdahl et al.~\cite{gravdahl2019robotic}
similarly integrates RRT with grasp-biased samples to find collision-free motions to a grasp goal.
Deng et al.~\cite{deng2018learning} sequence learned grasp planning into trajectory generation.
Pardi et al.~\cite{pardi2018choosing} propose selecting grasps that enable a motion planner to avoid obstacles.
Berenson et al.~\cite{berenson2011task} propose task space regions (TSR) for constraining robot motions.
GOMP integrates a variation of TSR constraints on the start and end configuration to compute an optimized motion that allows for degrees of freedom about pre-computed pick and place frames, and computing GOMP for multiple candidate grasps, allows one to find grasps that minimizes motion time while avoiding obstacles.
\section{Problem Definition}

Let $\vec{g}^{z-}$ be a top-down grasp produced by grasp analysis for a parallel-jaw gripper.
This grasp has a rotational degree of freedom associated with it based on the parallel-jaw gripper axis.
We note that this formulation can be extended to the rotational axis of a suction gripper's contact normal.
Let $\vec{q} \in \mathcal{C}$ be the complete specification of a robot's $n$ degrees of freedom (e.g., the angles of all $n$ joints in the robot), where $\mathcal{C} \subseteq \mathbb{R}^n$ is the set of all possible configurations (valid or not).
For robots with joint rotation limits, let $\vec{q}_{\mathrm{min}} \in \mathbb{R}^n$ and $\vec{q}_{\mathrm{max}} \in \mathbb{R}^n$ specify those limits.
Thus $\vec{q} \in \mathcal{C}$ implies that $\vec{q}_{\mathrm{min}} \le \vec{q} \le \vec{q}_{\mathrm{max}}$.
Let $\mathcal{C}_{\mathrm{obstacle}} \subset \mathcal{C}$ be the set of configurations that are in collision with an obstacle.
Let $\mathcal{C}_{\mathrm{free}} \subseteq \mathcal{C} \setminus \mathcal{C}_\mathrm{obstacle}$ be the set of configurations that are valid.
Let $p: \mathcal{C} \rightarrow \text{SE}(3)$ be the forward kinematic function for a robot.
The start configuration for a motion is a grasp $\vec{g}_{\mathrm{start}}^{z-} \in \text{SE}(3)$ produced by grasp analysis and corresponding to a top-down grasp. By adding a 1 degree of freedom in rotation $R_\vec{a}(\cdot)$ about axis $\vec{a}$ corresponding to the vector between the two grasp contact points, we get a set of starting grasps,
\[
G_\mathrm{start} = \left\{ 
\vec{g}_i
\;\middle|\;
\vec{g}_i = R_{\mathrm{a}}(\theta) \vec{g}_{\mathrm{start}}^{z-}, \;
\theta \in \left[-\frac{\pi}{2},\frac{\pi}{2}\right]
\right\},
\]
where here we limit the rotation to $\pm \frac{\pi}{2}$.
We note that the rotation limit may be set based on application needs.
We define $G_{\mathrm{goal}}$ similarly based on the target placement frame of the grasped object, though we note that the goal may have a different axis for its rotational degree of freedom, or may have no constraints on its rotation.

Let $\vec{v}_{\mathrm{max}} \in \mathbb{R}^n_+$ be the maximum velocity of each joint, and $\vec{v}_\mathrm{min} = -\vec{v}_{\mathrm{max}}$ be each joints minimum velocity.
Similarly, let $\vec{a}_{\mathrm{max}} \in \mathbb{R}^n_+$ be the maximum acceleration of each joint, and $\vec{a}_{\mathrm{min}} = -\vec{a}_{\mathrm{max}}$.
Let $\mathbf{\tau} \in \Tau$ be a sequence of joint configurations
$\mathbf{\tau} = (\vec{q}_0, \vec{q}_1, \ldots, \vec{q}_{\mathrm{end}})$, and
$f:\Tau \rightarrow \mathbb{R}_+$ be the time required to traverse the $\tau$ from $\vec{q}_0$ to $\vec{q}_{\mathrm{end}}$.

The objective of GOMP is to compute:
\begin{IEEEeqnarray}{rll}
\argmin_\tau \quad & f(\tau), \label{eq:tmin} \\
\textrm{s.t.} \quad & \vec{q}_{\mathrm{min}} \le \vec{q}_t \le \vec{q}_{\mathrm{max}}, & \quad \forall t \in [0,f(\tau)] \label{eq:qlimit} \\
& \vec{v}_{\mathrm{min}} \le \vec{v}_t \le \vec{v}_{\mathrm{max}}, & \quad \forall t \in [0,f(\tau)] \label{eq:vlimit} \\
& \vec{a}_{\mathrm{min}} \le \vec{a}_t \le \vec{a}_{\mathrm{max}}, & \quad \forall t \in [0,f(\tau)] \label{eq:alimit} \\
& \vec{q}_t \in \mathcal{C}_{\mathrm{free}}, & \quad \forall t \in [0,f(\tau)] \label{eq:obs} \\
& p(\vec{q}_0) \in G_{\mathrm{start}} \label{eq:q0} \\
& p(\vec{q}_\mathrm{end}) \in G_{\mathrm{goal}} \label{eq:qH}
\end{IEEEeqnarray}
where $\vec{q}_t$, $\vec{v}_t$, and $\vec{a}_t$ are the configuration, velocity, and accelerations at time $t$ respectively.
Thus, equations (\ref{eq:qlimit}), (\ref{eq:vlimit}), and (\ref{eq:alimit}) constrain the motion to the robot's mechanical limits,
equation (\ref{eq:obs}) constrains the motion to avoid obstacles, and 
equations (\ref{eq:q0}) and (\ref{eq:qH}) constrain the motion to start at a pick location, and end at a placement location.

\section{GOMP Algorithm}

\begin{algorithm}[t]
\caption{Grasp-Optimized Motion Planning}
\label{alg:ssqp}
\begin{algorithmic}[1]
\REQUIRE $\vec{g}_\mathrm{start}^{z-}$ and $\vec{g}_\mathrm{goal}^{z-}$ are start and goal gripper poses, \\
$H$ and $t_\mathrm{step}$ are the discretization parameters, \\
$\vec{q}_{\mathrm{min}}$, $\vec{q}_{\mathrm{max}}$, $\vec{v}_\mathrm{max}$, and
$\vec{a}_{\mathrm{max}}$ are the mechanical limits, \\
$\mat{D}$ is the collision model (depth map)
\STATE $\vec{q}_0 \leftarrow p^{-1}(\vec{g}_\mathrm{start}^{z-})$
\STATE $\vec{q}_H \leftarrow p^{-1}(\vec{g}_\mathrm{goal}^{z-})$
\STATE $\vec{x}_{H+1} \leftarrow \varnothing$
\STATE $\vec{x}_{H} \leftarrow $ spline interpolation from $\vec{q}_0$ to $\vec{q}_H$
\STATE $\mat{P} \leftarrow $ initialize matrix
\STATE $\mat{A}, \vec{b} \leftarrow $ linearize and initialize constraints from \\ \hfill $\vec{q}_\mathrm{min}$, $\vec{q}_\mathrm{max}$, $\vec{v}_\mathrm{max}$, $\vec{a}_\mathrm{max}$, $\vec{g}_\mathrm{start}^{z-}$, $\vec{g}_\mathrm{goal}^{z-}$, $\mat{D}$
\FOR{$h \leftarrow H$ down to  $2$}
  \FOR{$i \leftarrow 1, 2, \ldots$}
    \STATE $\vec{x}_h \leftarrow \argmin_\vec{x} \frac{1}{2} \vec{x}^T \mat{P} \vec{x}, \; \text{s.t. } \mat{A} \vec{x} \le \vec{b}$
    \IF{QP is infeasible \OR $i > i_\mathrm{max}$}
      \RETURN $\vec{x}_{h+1}$
    \ENDIF
    \STATE $\mat{A}, \vec{b} \leftarrow $ update linearization and trust region from \\ \hfill $\vec{x}_h[\vec{q}_0]$, $\vec{x}_h[\vec{q}_H]$, $\mat{D}$
    \IF{constraints and trust region in tolerance}
      \STATE \textbf{break}
    \ENDIF
  \ENDFOR
  \STATE $\vec{x}_{h-1} \leftarrow $ interpolate $\vec{x}_{h}$ to fit shorter trajectory
  \STATE $\mat{A}, \vec{b} \leftarrow $ add constraint to reflect $\vec{v}_h = 0$
\ENDFOR
\RETURN $\vec{x}_2$
\end{algorithmic}
\end{algorithm}

In this section we present a method for computing the constrained minimum-time trajectory of equation (\ref{eq:tmin}) using sequential quadratic programming (SQP).
The SQP method solves a sequence of quadratic program (QP) subproblems by establishing constraints based on trust regions.
We propose using QP solvers because fast implementations are readily available~(e.g., \cite{osqp}), and when solving a sequence of QP subproblems, each QP can be warm started using the solution to the previous subproblem.

A QP solver minimizes a quadratic objective in the form:
\begin{align*}
\argmin_{\vec{x}} & \; \frac{1}{2} \vec{x}^T \mat{P} \vec{x} + \vec{p}^T \vec{x} \\
\text{s.t.} &  \; \mat{A} \vec{x} \le \vec{b},
\end{align*}
where $\mat{P} \in \mathbf{S}_+$ is a positive semidefinite matrix, $\mat{A}$, and $\vec{b}$ represent the linear (and linearized) constraints.
For completeness, we include $\vec{p}^T$, but set it to $0$ in this paper.

To set up the QP, we first recast the problem so that $\tau$ is discretized over a finite horizon of $H$ steps, with each step separated by a fixed time interval $t_\mathrm{step}$.
We define $\vec{x}$ to match this definition as:
\[
\vec{x} = \begin{bmatrix}
\vec{q}_0^T & 
\vec{q}_1^T &
\cdots &
\vec{q}_H^T & %
\vec{v}_0^T & %
\vec{v}_1^T & %
\cdots &
\vec{v}_{H}
\end{bmatrix}^T,
\]
thus containing the configuration and velocity of each joint at each waypoint.

The recasting to a discretized fixed-interval sequence of waypoints serves several purposes:
the objective is now compatible with a QP solver, the constrained dynamics can be readily specified as linear constraints, and $t_\mathrm{step}$ can be set to match the real-type cycle frequency of the robot's control loop.
In section~\ref{sec:tmin}, this SQP formulation is then converted to a time-minimization.

The overall algorithm is presented in Alg.~\ref{alg:ssqp}.  The details of the algorithm are expanded on in the subsections that follow.

\subsection{Smooth trajectory objective}

After discretizing the trajectory, the next step is to set up the minimization objective for the QP.
Since the discretization means the trajectory has a fixed time interval, we instead have the QP encourage smooth trajectories by minimizing a the sum-of-squared accelerations as approximated by the difference in velocities.  Thus,
\[
    \mat{P} = \begin{bmatrix}
    0 & 0 \\
    0 & \mat{P}_v
    \end{bmatrix},
\]
where, $\mat{P}_v$ is a $(H+1) \times (H+1)$ matrix in the form:
\begin{equation}
    \mat{P}_v = \begin{bmatrix}
     2 & -1 & 0  &     0  & \cdots \\
    -1 &  2 & -1 &     0  &  \\
     0 & -1 &  2 &     -1 &  \\
\vdots &    & \ddots & \ddots &  \ddots \\
    \end{bmatrix}.
\end{equation}

\subsection{Dynamics constraints}

To enforce that the trajectory remains within the mechanical limits of the robot, we place linear constraints on $\vec{q}_i$, $\vec{v}_i$, and $\vec{a}_i$ to match the problem definition.
The position and velocity constraints can be directly applied to the coefficients of $\vec{x}$.
The accelerations constraint are set to
\[
-\vec{a}_\mathrm{max} \le \frac{1}{t_\mathrm{step}}(\vec{v}_{i+1} - \vec{v}_i) \le \vec{a}_\mathrm{max}.
\]

To enforce that the waypoints follow the dynamics implied by the velocity, we apply linear constraints using a \JI{\st{standard MPC}common model-predictive control} technique.  These constraints take the form:
\[ %
\vec{q}_{i+1} = \vec{q}_i + \vec{v}_i t_\mathrm{step}. %
\] %
With these constraints in place, the QP will enforce that the trajectory follows the dynamics implied by each waypoint while not exceeding the robot's mechanical limits.

Finally, to insure that the trajectory starts and ends with zero velocity, we add the constraints:
\[
\vec{v}_0 = 0 \qquad \vec{v}_H = 0
\]

\subsection{Obstacles}

To enable obstacle avoidance, we add a linearization of the constraints based on the robot's forward kinematics and the obstacle.  The method we apply follows directly from that of Trajopt~\cite{schulman2013finding}; we establish trust regions as linearized constraints in a QP.
In implementation, we observe that pick-and-place robots can operate with a simplified task-specific obstacle model---specifically a depth map.
As such, we can avoid the complexities associated with the GJK~\cite{gilbert1988fast} and EPA~\cite{cameron1986determining} methods for collision detection and instead incorporate a constraint based on each waypoint's position over the depth map.
We observe that an interior point optimization using boundary functions~\cite{kuntz2017fast} would also serve to avoid obstacles.

Assuming we define the positive $z$-axis as the up vector, obstacle avoidance constraint takes the form:
\[
z_{\mathrm{obstacle}} - p(\vec{q}_i^{(k)}) + \vec{J}_z^{(k)} \vec{q}_i^{(k)} \le
\vec{J}^{(k)}_z \vec{q}_i^{(k+1)},
\]
where, $z_\mathrm{obstacle}$ is the height of the obstacle that waypoint $i$ must be above, $\vec{J}_z^k$ is the $z$-axis translation row of the Jacobian of robot arm after SQP iteration $k$. 

\subsection{Start and goal constraints}

The pick-and-place operation has a set of start grasps $G_{\mathrm{start}}$ and a set of goal positions $G_{\mathrm{goal}}$ for the gripper.
To enforce the constraint that the trajectory starts with $p(\vec{q}_0) \in G_{\mathrm{start}}$ and ends at $p(\vec{q}_H) \in G_{\mathrm{goal}}$, we start with an initial QP constraint of having the $\vec{q}_0 = p^{-1}(\vec{g}_{\mathrm{start}}^{z-})$ and $\vec{q}_H = p^{-1}(\vec{g}_{\mathrm{goal}}^{z-})$, where $p^{-1}(\cdot)$ produces an inverse kinematic solution.
As described in the problem definition, these positions have at least one degree of freedom in the form of a rotation about the axis of the grasp.

The inequality constraint on the start configuration $\vec{q}_0$ that allows the rotation about a free axis is:
\[
\vec{b}_0^{(k+1)} - \bm \epsilon_0 \le
R_0 \vec{J}^{(k)} \vec{q}_0^{(k+1)} \le
\vec{b}_0^{(k+1)} + \bm \epsilon_0,
\]
where $R_0$ rotates the coordinate frame so that a single component of the Jacobian cooresponds to the degree of freedom.
Thus, $\bm \epsilon_0$ is a vector in which coefficient corresponding to that degree of freedom is large, and the remaining values in the vector are small.
The bounds of the inequality for the $k+1$-th QP is based on the solution to the $k$-th QP,
\[
\vec{b}_0^{(k+1)} = R_0 \left( 
\vec{g}_\mathrm{start}^{z-} - p(\vec{q}_0^{(k)}) +
\vec{J}^{(k)} \vec{q}_0^{(k)}
\right).
\]

The constraint on the goal configuration follows a similar formulation as the bounds on the start configuration, with potentially a different set of degrees of freedom.

\subsection{Time minimization}
\label{sec:tmin}

After solving the SQP with the constraints specified as above, $\vec{x}$ will correspond to a trajectory that satisfies dynamic constraints, avoids obstacles, and minimizes the sum-of-squared accelerations between waypoints.  To convert this formulation into a time minimization, we next repeatedly solve the SQP, reducing $H$ until the SQP is infeasible~\cite{osqp-infeasibility}, and taking the $\vec{x}$ associated with the smallest value of $H$ for which the SQP is feasible.
By finding the minimum $H$ we explicitly minimize the number of discrete time steps, and thus the overall time for the trajectory.

To implement the reduced horizon $H$, we convert the velocity constraint on the last waypoint from the inequality bound by $\vec{v}_{\mathrm{max}}$ to the equality constraint $\vec{v}_H = 0$.

We note that this process can be stopped at any time after the solution with the first feasible $H$, in which case the shortest time solution found so far would be usable as a trajectory for the robot.

\subsection{Warm starting}

The convergence time of QP solvers can benefit from providing a good initial estimate of the solution or \emph{warm starting} the solver~\cite{osqp}.
To speed up the SQP, we warm start several phases of the optimization process.

We warm start the first QP with a spline that interpolates between $\vec{q}_0$ and $\vec{q}_H$, and starting and stopping with zero velocity without regarding obstacles.

When reducing $H$ to find a minimum-time feasible trajectory, the solution from the previous optimization does not align with the optimization of the subsequent optimization.  %
As such we interpolate the solution from the previous $H$ to smaller set of waypoints.

\section{Results}

\begin{figure}
    \centering
    {\includegraphics[height=112pt]{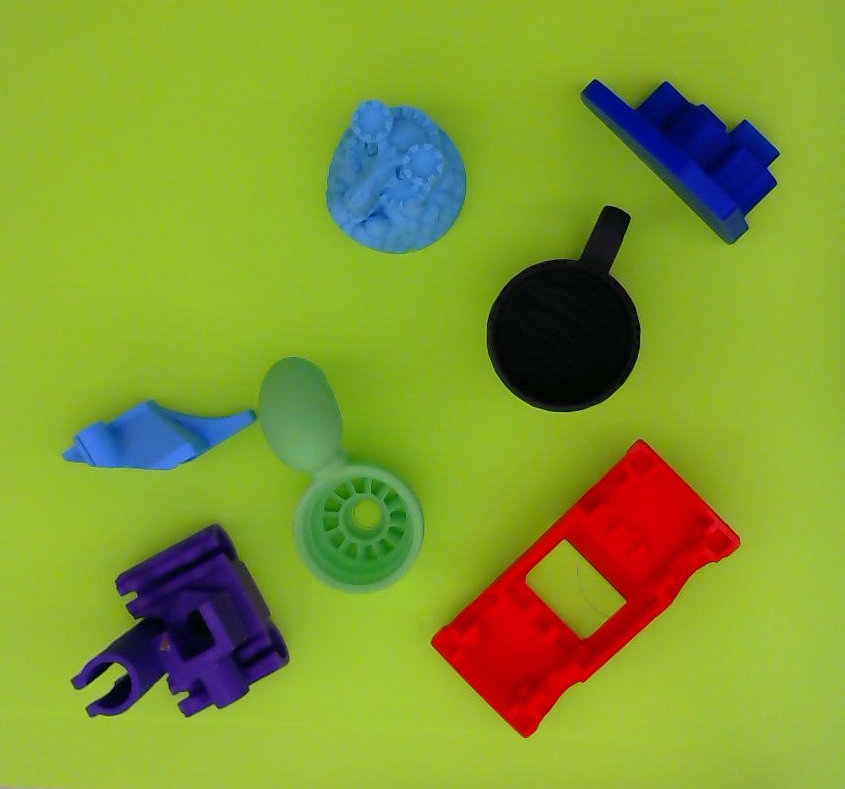}}\hfill%
    {\includegraphics[height=112pt]{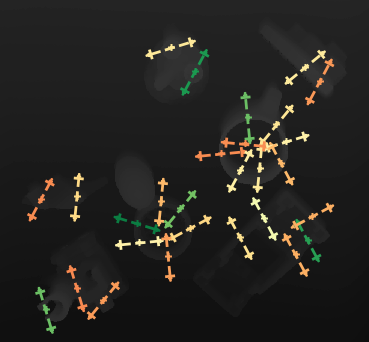}}
    \caption{In experimental setup, we take a top-down image of the objects to grasp (left) using a high-resolution depth camera.
    We then feed the image through a modified version of Dex-Net 4.0, which produces a diverse set of 28 candidate grasps (right).
    From these grasps, we then compute pick-to-place trajectories.}
    \label{fig:grasp_analysis}
\end{figure}

\newcommand{\videostill}[1]{\subfloat[]{\includegraphics[width=79pt,trim={368 0 290 68},clip]{#1}}}

\begin{figure*}
\centering
\begin{tabular}{@{}c@{\hspace{6pt}}c@{\hspace{6pt}}c@{\hspace{6pt}}c@{\hspace{6pt}}c@{\hspace{6pt}}c@{}}
  \videostill{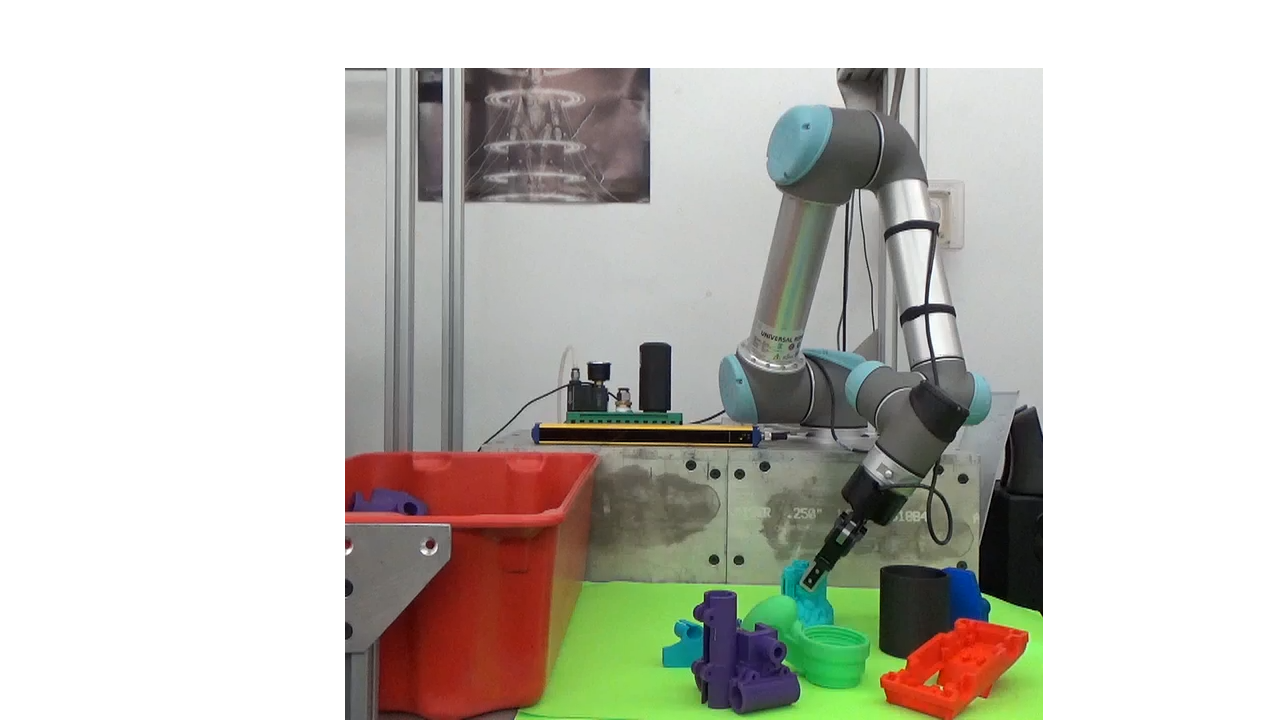}&
  \videostill{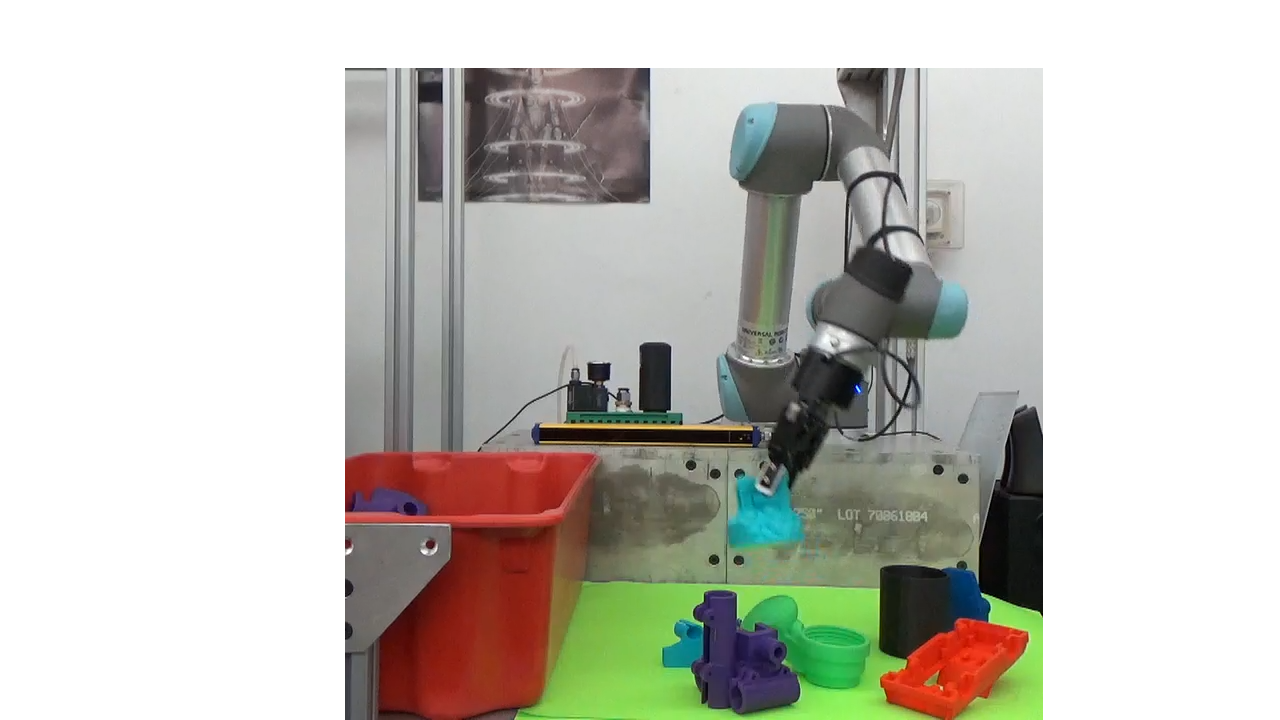}&
  \videostill{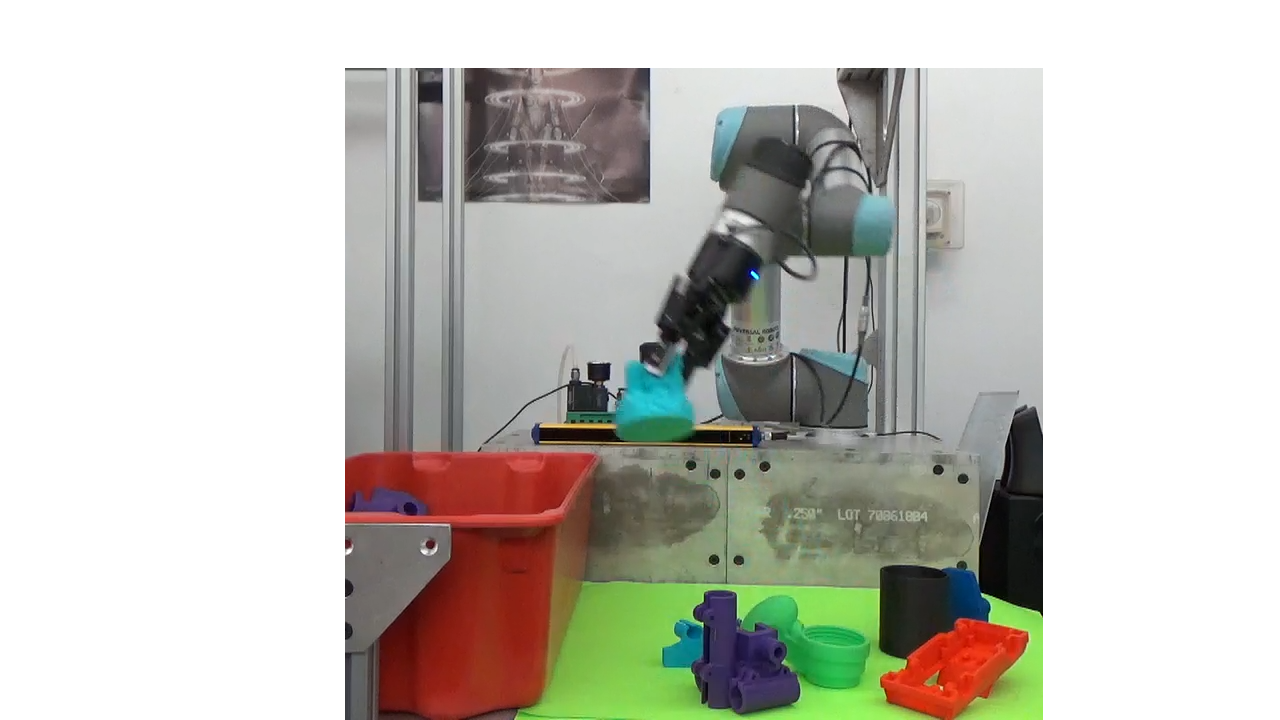}&
  \videostill{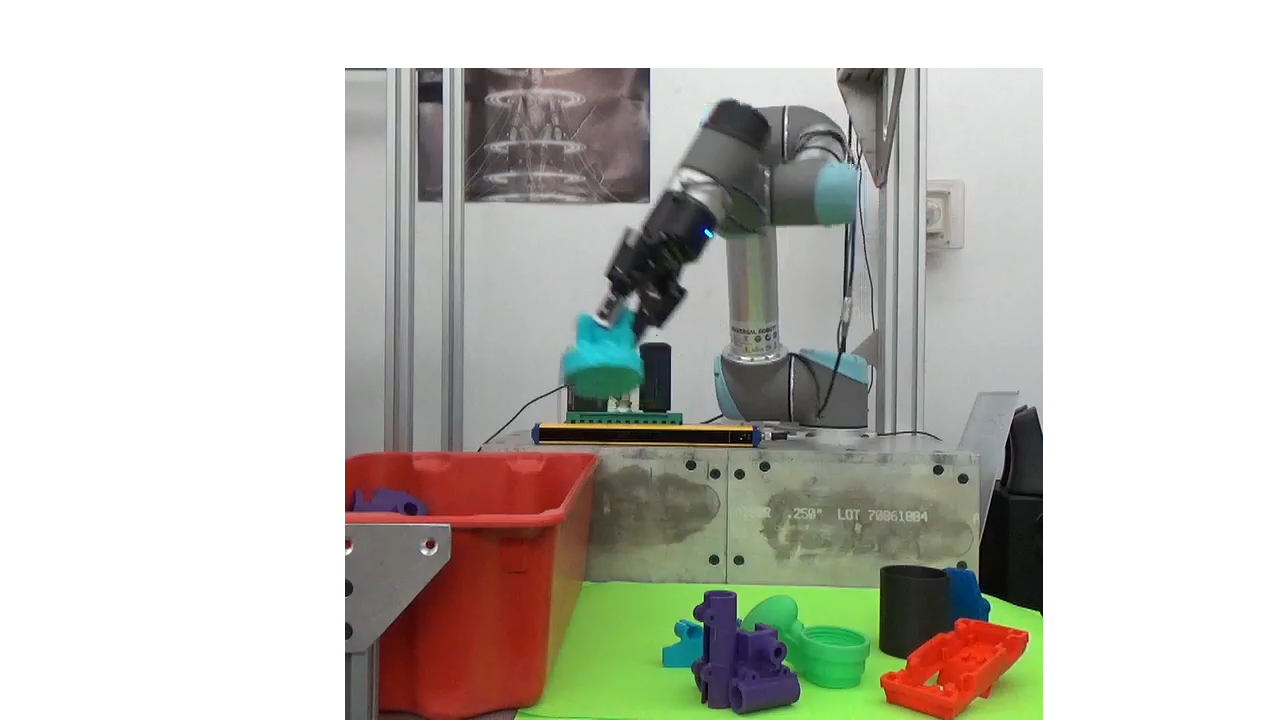}&
  \videostill{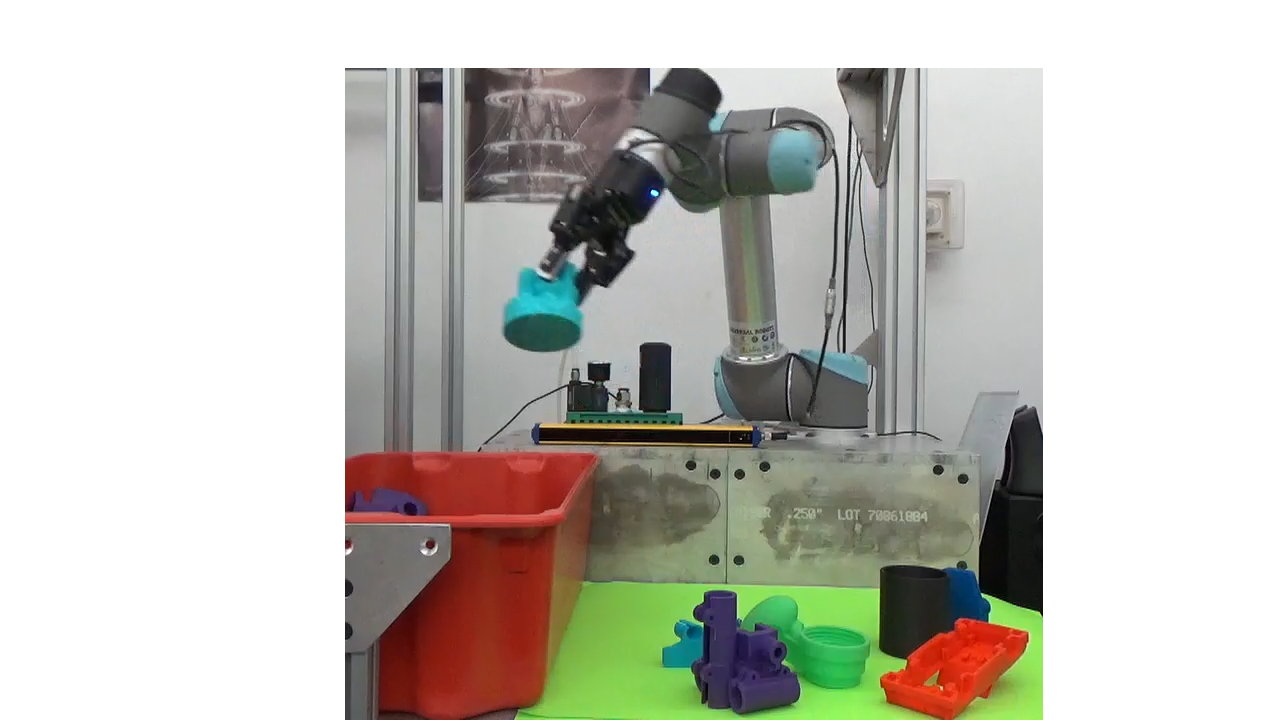}& 
  \videostill{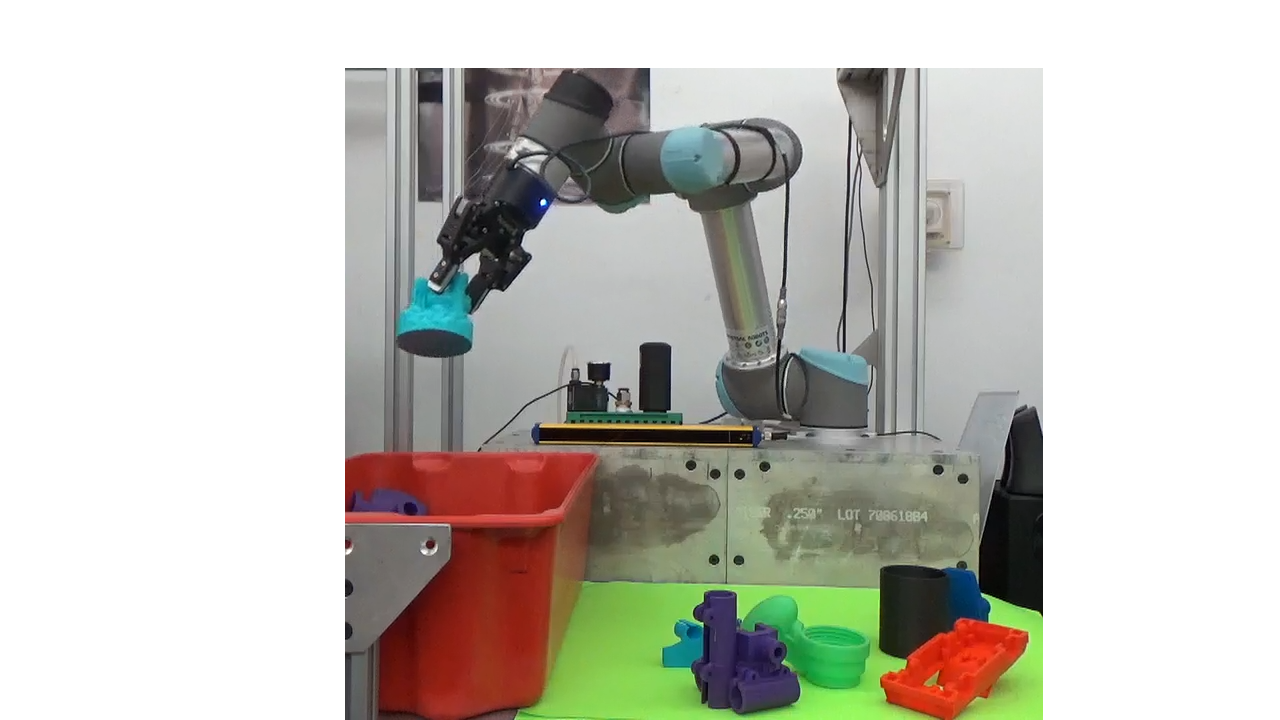}
\end{tabular}
\caption{UR5 robot moves quickly to transport picked object to placement over bin.  The UR5 moves within its joint limits, and with both pick and placement orientation optimized as part of the motion planning process.  This motion takes 0.43~seconds.}
\label{fig:ur5_physical}
\end{figure*}

\begin{figure*}
  \centering
  \begin{tikzpicture}
    \node[inner sep=0pt, outer sep=0pt] (plot) at (0,0) {\input{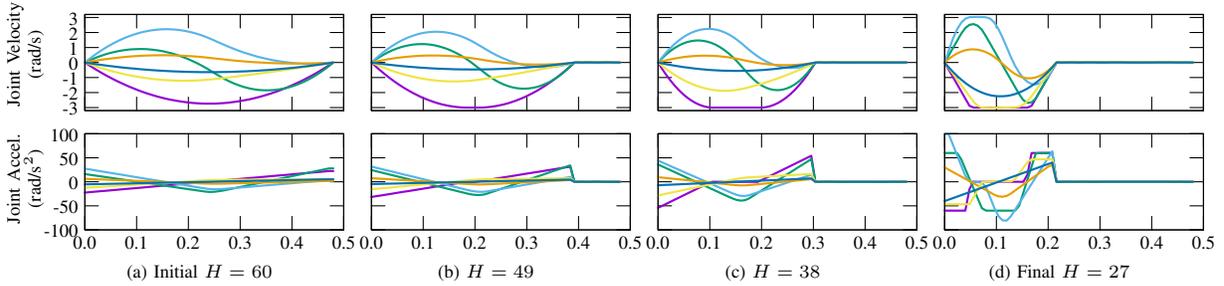}};
    \node[inner sep=0pt, outer sep=0pt, left=0pt of plot, yshift=0.4in] {%
      \rotatebox[origin=c]{90}{\parbox{0.6in}{\centering \scriptsize Joint Velocity \\ (rad/s)}}};
     \node[inner sep=0pt, outer sep=0pt, left=0pt of plot, yshift=-0.25in] {%
       \rotatebox[origin=c]{90}{\parbox{0.6in}{\centering \scriptsize Joint Accel. \\ (rad/s$^2$)}}};
     \node[inner sep=0pt, outer sep=0pt, below=6pt of plot, xshift=-2.25in] {%
       \scriptsize (a) Initial $H=60$};
     \node[inner sep=0pt, outer sep=0pt, below=6pt of plot, xshift=-0.75in] {%
       \scriptsize (b) $H=49$};
     \node[inner sep=0pt, outer sep=0pt, below=6pt of plot, xshift=0.75in] {%
       \scriptsize (c) $H=38$};
     \node[inner sep=0pt, outer sep=0pt, below=6pt of plot, xshift=2.25in] {%
       \scriptsize (d) Final $H=27$};
  \end{tikzpicture}
  \caption{\textbf{Joint velocities and accelerations computed during time optimization of the trajectory.}
    In these plots, the velocity and acceleration of each of the 6 joints on the UR5 robot is plotted in a separate color.  Each column represents the result of trajectory optimization for a successively smaller value of $H$.
  The time optimization of the trajectory starts with a value of $H$ and subsequently shrinks this value by constraining the velocity to zero at the end of the trajectory.  Trajectory optimization stops once the QP is infeasible.  In this case, the shortest time is $H = 27$, which, at $t_{\mathrm{step}} = 0.008$, results in a trajectory that requires 0.216 seconds to execute.}
  \label{fig:joint_trajectories}
\end{figure*}

To test the performance and compatibility of the trajectories GOMP computes, we apply them to a physical setup resembling what one may find at a pick-and-place station in a warehouse.
In this setup, a high-resolution depth camera produces an image of the objects to grasp (Fig.~\ref{fig:grasp_analysis}~(a)).
This image is then sent to our modified version of Dex-Net 4.0 to generate a diverse set of grasp candidate poses (Fig.~\ref{fig:grasp_analysis}~(b)).
From the grasp candidates, we compute a trajectory for the UR5 robot that takes the robot from pick-point to placement, while avoiding obstacles including the side of the placement bin.
This physical setup, along with the execution of an example trajectory, is shown in Fig.~\ref{fig:ur5_physical}.

We modify the original Dex-Net 4.0 to produce a diverse set of grasp candidates for GOMP by changing the sampling method from argmax to rejection-sampling with a minimum distance constraint between grasps in image space.

In the physical experiment, each grasp is initially computed as a top-down grasp.
We set the SQP constraints to allow rotation about the grasp point by up to 45 degrees in either direction.
In Fig.~\ref{fig:ur5_physical}~(a), the SQP computed an initial grasp that aligns to the computed grasp points, but is rotated to allow for a faster motion from pick-point to placement.

At any point in time after the initial computation of a trajectory, the robot has a valid trajectory.
With additional compute time, successive reduction of $H$ allows for a faster trajectory.
\JI{An example of \st{T}t}his process of computing faster trajectories is shown in Fig.~\ref{fig:joint_trajectories}.
In this figure, we can see an initial smooth trajectory that is successively made faster until it cannot be shortened any further due to mechanical limits of the robot.

\begin{table}[t]
    \centering
    \caption{Timing comparison for 28 pick-to-place motions}
    \begin{tabular}{r|rrr}
         & baseline & GOMP & speedup\\
         \hline
    mean & \SI{5.042}{\second} & \SI{0.544}{\second} & 9.2$\times$ \\
    stdev & \SI{0.440}{\second} & \SI{0.172}{\second} & \\
    min & \SI{4.144}{\second} & \SI{0.256}{\second} & 16.2$\times$ \\
    max & \SI{5.624}{\second} & \SI{0.976}{\second} & 5.8$\times$
    \end{tabular}
    \label{tab:timing}
\end{table}

We compare the motion execution time for the 28 grasps shown in Fig.~\ref{fig:ur5_physical}, to a baseline trajectory that uses top-down grasps and performs 3 steps: lift to safe height, move over bin, then lower to placement height.
The baseline motion does not incorporate obstacle avoidance, instead, it avoids obstacles by lifting the object sufficiently high to avoid foreseeable collisions in the workspace.  This baseline is run at the UR5's default speed.
The comparison between the baseline and the optimized motions are shown in table~\ref{tab:timing}.
From the table, we observe that GOMP performs faster in all tested cases, and on average provides speedup that is 9.2 faster than the baseline.
These savings can be passed on to the complete grasping pipeline to speed up the pick and place process.

\begin{table}[t]
    \caption{Per-object GOMP motion times}
    \centering
    \begin{tabular}{r|rrrr}
         object (color) & grasps & minimum & mean & maximum \\
         \hline
         Car (orange) & 5 & \textbf{\SI{0.400}{\second}} & \SI{0.659}{\second} & \SI{0.736}{\second} \\
         Castle (blue, right) & 2 & \textbf{\SI{0.432}{\second}} & \SI{0.464}{\second} & \SI{0.496}{\second} \\
         Clamp (dark blue) & 2 & \textbf{\SI{0.480}{\second}} & \SI{0.512}{\second} & \SI{0.544}{\second} \\
         Mug (black) & 8 & \textbf{\SI{0.382}{\second}} & \SI{0.578}{\second} & \SI{0.976}{\second} \\
         Nozzle (blue, left) & 2 & \textbf{\SI{0.400}{\second}} & \SI{0.440}{\second} & \SI{0.480}{\second} \\
         Pipe connector (purple) & 3 & \textbf{\SI{0.384}{\second}} & \SI{0.501}{\second} & \SI{0.384}{\second} \\
         Turbine (green) & 6 & \textbf{\SI{0.256}{\second}} & \SI{0.496}{\second} & \SI{0.896}{\second} \\
    \end{tabular}
    \label{tab:min_grasp}
\end{table}

We also test the ability of the trajectory optimization to be integrated with grasp selection.
In Fig.~\ref{fig:grasp_analysis}, each object has multiple candidate grasp points.
For each object, we compute trajectories to all of its candidate grasp points in parallel, and then select the grasp that results in the shortest execution time.
In table~\ref{tab:min_grasp} we observe that selecting the grasp that produces the shortest motion can sometime produce an additional performance benefit.

The bin-picking pipeline in these experiments requires the following steps:
(a) imaging ($1.200\pm0.051$~s),
(b) grasp analysis ($0.529\pm0.041$~s)
(c) motion to grasp,
(d) gripper closing ($0.280\pm0.140$~s),
(e) motion to placement, and
(f) gripper opening ($0.280\pm0.140$~s).
Plugging the mean values for baseline motions and GOMP motions from table~\ref{tab:timing}, the mean total pick time for baseline motions is 12.373~s, and with GOMP motions is 3.377~s.
This suggests that GOMP can speed up our pipeline's mean total pick time by 3.6$\times$. %
We caveat that this PPH has not been confirmed with a fully \JI{\st{implemented}integrated} system.

\section{Conclusion}

We presented a time-optimizing motion planner that speeds up pick-and-place operation by incorporating joint limits, obstacle avoidance, and allowing a degree of freedom to be incorporated into grasping and placement positions.
By first converting the problem from a time-minimization problem to minimization that encourages smooth velocities and linearizing constraints, we are able to recast the problem to a sequential quadratic program (SQP).
After solving the SQP, we then repeatedly shorten the trajectory's time horizon in the SQP---stopping at any time, or when the SQP is detected as infeasible, resulting in a fast motion plan.
When comparing to a baseline motion, the optimized motion plan runs significantly faster, creating an opportunity to speed up the entire pick-and-place operation.

In future work, we plan to test this motion planner on different robots with differing degrees of freedom, as well as integrate it into the Dex-Net pipeline.
We also see opportunities to extend this work to account for additional constraints and trajectory objectives that one might encounter when deploying this to different robots and environments---including, for example, generating minimum-jerk trajectories, and avoiding dynamic obstacles.

\section*{Acknowledgments}
\footnotesize
This research was performed at the AUTOLAB at UC Berkeley in affiliation with the Berkeley AI Research (BAIR) Lab, Berkeley Deep Drive (BDD), the Real-Time Intelligent Secure Execution (RISE) Lab, and the CITRIS ``People and Robots'' (CPAR) Initiative. Authors were also supported by the Scalable Collaborative Human-Robot Learning (SCHooL) Project, a NSF National Robotics Initiative Award 1734633, and in part by donations from Google and Toyota Research Institute.  We thank our colleagues who provided helpful feedback and suggestions, in particular Ashwin Balakrishna and Brijen Thananjeyan.  This article solely reflects the opinions and conclusions of its authors and do not reflect the views of the sponsors or their associated entities.  

\bibliographystyle{IEEEtran}
\bibliography{IEEEabrv,references}

\begin{thebibliography}{10}
\providecommand{\url}[1]{#1}
\csname url@rmstyle\endcsname
\providecommand{\newblock}{\relax}
\providecommand{\bibinfo}[2]{#2}
\providecommand\BIBentrySTDinterwordspacing{\spaceskip=0pt\relax}
\providecommand\BIBentryALTinterwordstretchfactor{4}
\providecommand\BIBentryALTinterwordspacing{\spaceskip=\fontdimen2\font plus
\BIBentryALTinterwordstretchfactor\fontdimen3\font minus
  \fontdimen4\font\relax}
\providecommand\BIBforeignlanguage[2]{{%
\expandafter\ifx\csname l@#1\endcsname\relax
\typeout{** WARNING: IEEEtran.bst: No hyphenation pattern has been}%
\typeout{** loaded for the language `#1'. Using the pattern for}%
\typeout{** the default language instead.}%
\else
\language=\csname l@#1\endcsname
\fi
#2}}

\bibitem{morrison2019learning}
D.~Morrison, P.~Corke, and J.~Leitner, ``Learning robust, real-time, reactive
  robotic grasping,'' \emph{The International Journal of Robotics Research}, p.
  0278364919859066, 2019.

\bibitem{satish2019policy}
V.~Satish, J.~Mahler, and K.~Goldberg, ``On-policy dataset synthesis for
  learning robot grasping policies using fully convolutional deep networks,''
  \emph{IEEE Robotics \& Automation Letters}, vol.~4, no.~2, pp. 1357--1364,
  2019.

\bibitem{UR5robot}
\BIBentryALTinterwordspacing
U.~Robotics. {UR5} collaborative robot arm. [Online]. Available:
  \url{https://web.archive.org/web/20190815054522/https://www.universal-robots.com/products/ur5-robot/}
\BIBentrySTDinterwordspacing

\bibitem{Robotiq2f85}
\BIBentryALTinterwordspacing
Robotiq. {2F-85} and {2F-140} grippers. [Online]. Available:
  \url{https://web.archive.org/web/20190519030456/https://robotiq.com/products/2f85-140-adaptive-robot-gripper}
\BIBentrySTDinterwordspacing

\bibitem{mahler2019learning}
J.~Mahler, M.~Matl, V.~Satish, M.~Danielczuk, B.~DeRose, S.~McKinley, and
  K.~Goldberg, ``Learning ambidextrous robot grasping policies,'' \emph{Science
  Robotics}, vol.~4, no.~26, 2019.

\bibitem{mahler2016dex}
J.~Mahler, F.~T. Pokorny, B.~Hou, M.~Roderick, M.~Laskey, M.~Aubry,
  K.~Kohlhoff, T.~Kr{\"o}ger, J.~Kuffner, and K.~Goldberg, ``Dex-net 1.0: A
  cloud-based network of 3d objects for robust grasp planning using a
  multi-armed bandit model with correlated rewards,'' in \emph{{Proc. {IEEE}
  Int. Conf. Robotics and Automation (ICRA)}}.\hskip 1em plus 0.5em minus
  0.4em\relax IEEE, 2016, pp. 1957--1964.

\bibitem{mahler2017dex}
J.~Mahler, J.~Liang, S.~Niyaz, M.~Laskey, R.~Doan, X.~Liu, J.~Aparicio, and
  K.~Goldberg, ``Dex-net 2.0: Deep learning to plan robust grasps with
  synthetic point clouds and analytic grasp metrics,'' in \emph{Proc. Robotics:
  Science and Systems (RSS)}, 2017.

\bibitem{mahler2017learning}
J.~Mahler and K.~Goldberg, ``Learning deep policies for robot bin picking by
  simulating robust grasping sequences,'' in \emph{Conf. on Robot Learning
  (CoRL)}, 2017, pp. 515--524.

\bibitem{ten2017grasp}
A.~ten Pas, M.~Gualtieri, K.~Saenko, and R.~Platt, ``Grasp pose detection in
  point clouds,'' \emph{Int. Journal of Robotics Research (IJRR)}, vol.~36, no.
  13-14, pp. 1455--1473, 2017.

\bibitem{kappler2015leveraging}
D.~Kappler, J.~Bohg, and S.~Schaal, ``Leveraging big data for grasp planning,''
  in \emph{{Proc. {IEEE} Int. Conf. Robotics and Automation (ICRA)}}.\hskip 1em
  plus 0.5em minus 0.4em\relax IEEE, 2015, pp. 4304--4311.

\bibitem{lenz2015deep}
I.~Lenz, H.~Lee, and A.~Saxena, ``Deep learning for detecting robotic grasps,''
  \emph{Int. Journal of Robotics Research (IJRR)}, vol.~34, no. 4-5, pp.
  705--724, 2015.

\bibitem{johns2016deep}
E.~Johns, S.~Leutenegger, and A.~J. Davison, ``Deep learning a grasp function
  for grasping under gripper pose uncertainty,'' in \emph{Proc. IEEE/RSJ Int.
  Conf. on Intelligent Robots and Systems (IROS)}.\hskip 1em plus 0.5em minus
  0.4em\relax IEEE, 2016, pp. 4461--4468.

\bibitem{viereck2017learning}
U.~Viereck, A.~t. Pas, K.~Saenko, and R.~Platt, ``Learning a visuomotor
  controller for real world robotic grasping using simulated depth images,'' in
  \emph{Conf. on Robot Learning (CoRL)}, 2017.

\bibitem{kalashnikov2018scalable}
D.~Kalashnikov, A.~Irpan, P.~Pastor, J.~Ibarz, A.~Herzog, E.~Jang, D.~Quillen,
  E.~Holly, M.~Kalakrishnan, V.~Vanhoucke, and S.~Levine, ``Scalable deep
  reinforcement learning for vision-based robotic manipulation,'' in
  \emph{Conf. on Robot Learning (CoRL)}, 2018, pp. 651--673.

\bibitem{pinto2016supersizing}
L.~Pinto and A.~Gupta, ``Supersizing self-supervision: Learning to grasp from
  50k tries and 700 robot hours,'' in \emph{{Proc. {IEEE} Int. Conf. Robotics
  and Automation (ICRA)}}.\hskip 1em plus 0.5em minus 0.4em\relax IEEE, 2016,
  pp. 3406--3413.

\bibitem{levine2016learning}
S.~Levine, P.~Pastor, A.~Krizhevsky, and D.~Quillen, ``Learning hand-eye
  coordination for robotic grasping with large-scale data collection.''\hskip
  1em plus 0.5em minus 0.4em\relax Springer, 2016, pp. 173--184.

\bibitem{bousmalis2018using}
K.~Bousmalis, A.~Irpan, P.~Wohlhart, Y.~Bai, M.~Kelcey, M.~Kalakrishnan,
  L.~Downs, J.~Ibarz, P.~Pastor, K.~Konolige, \emph{et~al.}, ``Using simulation
  and domain adaptation to improve efficiency of deep robotic grasping,'' in
  \emph{{Proc. {IEEE} Int. Conf. Robotics and Automation (ICRA)}}.\hskip 1em
  plus 0.5em minus 0.4em\relax IEEE, 2018, pp. 4243--4250.

\bibitem{mousavian20196}
A.~Mousavian, C.~Eppner, and D.~Fox, ``6-dof graspnet: Variational grasp
  generation for object manipulation,'' in \emph{Proc. {IEEE} Int. Conf. on
  Computer Vision (ICCV)}, 2019, pp. 2901--2910.

\bibitem{murali20196}
A.~Murali, A.~Mousavian, C.~Eppner, C.~Paxton, and D.~Fox, ``6-dof grasping for
  target-driven object manipulation in clutter,'' \emph{arXiv preprint
  arXiv:1912.03628}, 2019.

\bibitem{yan2018learning}
X.~Yan, J.~Hsu, M.~Khansari, Y.~Bai, A.~Pathak, A.~Gupta, J.~Davidson, and
  H.~Lee, ``Learning 6-dof grasping interaction via deep geometry-aware 3d
  representations,'' in \emph{{Proc. {IEEE} Int. Conf. Robotics and Automation
  (ICRA)}}.\hskip 1em plus 0.5em minus 0.4em\relax IEEE, 2018, pp. 1--9.

\bibitem{liu2019generating}
M.~Liu, Z.~Pan, K.~Xu, K.~Ganguly, and D.~Manocha, ``Generating grasp poses for
  a high-dof gripper using neural networks,'' in \emph{Proc. IEEE/RSJ Int.
  Conf. on Intelligent Robots and Systems (IROS)}, 2019.

\bibitem{eppner2019billion}
C.~Eppner, A.~Mousavian, and D.~Fox, ``A billion ways to grasp: An evaluation
  of grasp sampling schemes on a dense, physics-based grasp data set,'' in
  \emph{Int. S. Robotics Research (ISRR)}, 2019.

\bibitem{Kavraki1996_TRA}
L.~E. Kavraki, P.~Svestka, J.-C. Latombe, and M.~Overmars, ``{Probabilistic
  roadmaps for path planning in high dimensional configuration spaces},''
  \emph{IEEE Trans. Robotics and Automation}, vol.~12, no.~4, pp. 566--580,
  1996.

\bibitem{Karaman2011_IJRR}
S.~Karaman and E.~Frazzoli, ``{Sampling-based algorithms for optimal motion
  planning},'' \emph{Int. Journal of Robotics Research (IJRR)}, vol.~30, no.~7,
  pp. 846--894, June 2011.

\bibitem{kuffner2005efficient}
J.~Kuffner and S.~LaValle, ``An efficient approach to path planning using
  balanced bidirectional rrt search,'' \emph{Robotics Institute, Carnegie
  Mellon University, Pittsburgh, PA, Tech. Rep}, 2005.

\bibitem{Amato1999_ICRA}
N.~M. Amato and L.~K. Dale, ``{Probabilistic roadmap methods are embarrassingly
  parallel},'' in \emph{{Proc. {IEEE} Int. Conf. Robotics and Automation
  (ICRA)}}, May 1999, pp. 688--694.

\bibitem{Ichnowski2014_TRO}
J.~Ichnowski and R.~Alterovitz, ``Scalable multicore motion planning using
  lock-free concurrency,'' \emph{IEEE Trans. Robotics}, vol.~30, no.~5, pp.
  1123--1136, 2014.

\bibitem{pan2012collision}
J.~Pan, L.~Zhang, and D.~Manocha, ``Collision-free and smooth trajectory
  computation in cluttered environments,'' \emph{The International Journal of
  Robotics Research}, vol.~31, no.~10, pp. 1155--1175, 2012.

\bibitem{pham2014general}
Q.-C. Pham, ``A general, fast, and robust implementation of the time-optimal
  path parameterization algorithm,'' \emph{IEEE Transactions on Robotics},
  vol.~30, no.~6, pp. 1533--1540, 2014.

\bibitem{kunz2012time}
T.~Kunz and M.~Stilman, ``Time-optimal trajectory generation for path following
  with bounded acceleration and velocity,'' \emph{Robotics: Science and Systems
  VIII}, pp. 1--8, 2012.

\bibitem{schulman2013finding}
J.~Schulman, J.~Ho, A.~X. Lee, I.~Awwal, H.~Bradlow, and P.~Abbeel, ``Finding
  locally optimal, collision-free trajectories with sequential convex
  optimization.'' in \emph{Robotics: science and systems}, vol.~9, no.~1.\hskip
  1em plus 0.5em minus 0.4em\relax Citeseer, 2013, pp. 1--10.

\bibitem{ratliff2009chomp}
N.~Ratliff, M.~Zucker, J.~A. Bagnell, and S.~Srinivasa, ``{CHOMP}: Gradient
  optimization techniques for efficient motion planning,'' 2009.

\bibitem{kalakrishnan2011stomp}
M.~Kalakrishnan, S.~Chitta, E.~Theodorou, P.~Pastor, and S.~Schaal, ``{STOMP}:
  Stochastic trajectory optimization for motion planning,'' in \emph{2011 IEEE
  international conference on robotics and automation}.\hskip 1em plus 0.5em
  minus 0.4em\relax IEEE, 2011, pp. 4569--4574.

\bibitem{park2012itomp}
C.~Park, J.~Pan, and D.~Manocha, ``{ITOMP}: Incremental trajectory optimization
  for real-time replanning in dynamic environments,'' in \emph{Twenty-Second
  International Conference on Automated Planning and Scheduling}, 2012.

\bibitem{kuntz2017fast}
A.~Kuntz, C.~Bowen, and R.~Alterovitz, ``Fast anytime motion planning in point
  clouds by interleaving sampling and interior point optimization,''
  \emph{ISRR, 2017}, 2017.

\bibitem{campana2016gradient}
M.~Campana, F.~Lamiraux, and J.-P. Laumond, ``A gradient-based path
  optimization method for motion planning,'' \emph{Advanced Robotics}, vol.~30,
  no. 17-18, pp. 1126--1144, 2016.

\bibitem{mukadam2018continuous}
M.~Mukadam, J.~Dong, X.~Yan, F.~Dellaert, and B.~Boots, ``Continuous-time
  gaussian process motion planning via probabilistic inference,'' \emph{The
  International Journal of Robotics Research}, vol.~37, no.~11, pp. 1319--1340,
  2018.

\bibitem{dragan2011manipulation}
A.~D. Dragan, N.~D. Ratliff, and S.~S. Srinivasa, ``Manipulation planning with
  goal sets using constrained trajectory optimization,'' in \emph{2011 IEEE
  International Conference on Robotics and Automation}.\hskip 1em plus 0.5em
  minus 0.4em\relax IEEE, 2011, pp. 4582--4588.

\bibitem{wang2019manipulation}
L.~Wang, Y.~Xiang, and D.~Fox, ``Manipulation trajectory optimization with
  online grasp synthesis and selection,'' \emph{arXiv preprint
  arXiv:1911.10280}, 2019.

\bibitem{vahrenkamp2010integrated}
N.~Vahrenkamp, M.~Do, T.~Asfour, and R.~Dillmann, ``Integrated grasp and motion
  planning,'' in \emph{2010 IEEE International Conference on Robotics and
  Automation}.\hskip 1em plus 0.5em minus 0.4em\relax IEEE, 2010, pp.
  2883--2888.

\bibitem{fontanals2014integrated}
J.~Fontanals, B.-A. Dang-Vu, O.~Porges, J.~Rosell, and M.~A. Roa, ``Integrated
  grasp and motion planning using independent contact regions,'' in \emph{2014
  IEEE-RAS International Conference on Humanoid Robots}.\hskip 1em plus 0.5em
  minus 0.4em\relax IEEE, 2014, pp. 887--893.

\bibitem{gravdahl2019robotic}
I.~Gravdahl, K.~Seel, and E.~I. Gr{\o}tli, ``Robotic bin-picking under
  geometric end-effector constraints: Bin placement and grasp selection,'' in
  \emph{2019 7th International Conference on Control, Mechatronics and
  Automation (ICCMA)}.\hskip 1em plus 0.5em minus 0.4em\relax IEEE, 2019, pp.
  197--203.

\bibitem{deng2018learning}
Z.~Deng, X.~Zheng, L.~Zhang, and J.~Zhang, ``A learning framework for semantic
  reach-to-grasp tasks integrating machine learning and optimization,''
  \emph{Robotics and Autonomous Systems}, vol. 108, pp. 140--152, 2018.

\bibitem{pardi2018choosing}
T.~Pardi, R.~Stolkin, \emph{et~al.}, ``Choosing grasps to enable collision-free
  post-grasp manipulations,'' in \emph{2018 IEEE-RAS 18th International
  Conference on Humanoid Robots (Humanoids)}.\hskip 1em plus 0.5em minus
  0.4em\relax IEEE, 2018, pp. 299--305.

\bibitem{berenson2011task}
D.~Berenson, S.~Srinivasa, and J.~Kuffner, ``Task space regions: A framework
  for pose-constrained manipulation planning,'' \emph{The International Journal
  of Robotics Research}, vol.~30, no.~12, pp. 1435--1460, 2011.

\bibitem{osqp}
B.~Stellato, G.~Banjac, P.~Goulart, A.~Bemporad, and S.~Boyd, ``{OSQP}: An
  operator splitting solver for quadratic programs,'' \emph{ArXiv e-prints},
  Nov. 2017.

\bibitem{gilbert1988fast}
E.~G. Gilbert, D.~W. Johnson, and S.~S. Keerthi, ``A fast procedure for
  computing the distance between complex objects in three-dimensional space,''
  \emph{IEEE Journal on Robotics and Automation}, vol.~4, no.~2, pp. 193--203,
  1988.

\bibitem{cameron1986determining}
S.~Cameron and R.~Culley, ``Determining the minimum translational distance
  between two convex polyhedra,'' in \emph{Proceedings. 1986 IEEE International
  Conference on Robotics and Automation}, vol.~3.\hskip 1em plus 0.5em minus
  0.4em\relax IEEE, 1986, pp. 591--596.

\bibitem{osqp-infeasibility}
\BIBentryALTinterwordspacing
G.~Banjac, P.~Goulart, B.~Stellato, and S.~Boyd, ``Infeasibility detection in
  the alternating direction method of multipliers for convex optimization,''
  \emph{Journal of Optimization Theory and Applications}, vol. 183, no.~2, pp.
  490--519, 2019. [Online]. Available:
  \url{https://doi.org/10.1007/s10957-019-01575-y}
\BIBentrySTDinterwordspacing

\end{thebibliography}

\end{document}